\documentclass[10pt,twocolumn,letterpaper]{article}

\usepackage{cuted}    %
\usepackage{capt-of}  %
\usepackage[pagenumbers]{cvpr} %

\usepackage{booktabs}
\usepackage{multirow}
\usepackage{graphicx}
\usepackage[table]{xcolor} %
\usepackage{colortbl}
\usepackage[accsupp]{axessibility}  %
\definecolor{colgray}{gray}{0.95} %
\definecolor{orionblue}{RGB}{235, 245, 255} %
\definecolor{textgray}{gray}{0.5} %
\definecolor{oriongreen}{HTML}{2ca02c}

\definecolor{cvprblue}{rgb}{0.21,0.49,0.74}
\usepackage[pagebackref,breaklinks,colorlinks,allcolors=cvprblue]{hyperref}

\def\paperID{4} %
\def\confName{CVPR}
\def\confYear{2026}

\title{Orion-Lite: Distilling LLM Reasoning into Efficient Vision-Only Driving Models}

\author{
    Jing Gu \quad Niccol\`o Cavagnero \quad Gijs Dubbelman \\
    Eindhoven University of Technology \\
    {\tt\small \{j.gu1, n.cavagnero, g.Dubbelman\}@tue.nl} \\ %
    \url{https://github.com/tue-mps/Orion-Lite}
}

\begin{document}
\maketitle

\begin{abstract}
Leveraging the general world knowledge of Large Language Models (LLMs) holds significant promise for improving the ability of autonomous driving systems to handle rare and complex scenarios. While integrating LLMs into Vision-Language-Action (VLA) models has yielded state-of-the-art performance, their massive parameter counts pose severe challenges for latency-sensitive and energy-efficient deployment. Distilling LLM knowledge into a compact driving model offers a compelling solution to retain these reasoning capabilities while maintaining a manageable computational footprint. Although previous works have demonstrated the efficacy of distillation, these efforts have primarily focused on relatively simple scenarios and open-loop evaluations. Therefore, in this work, we investigate LLM distillation in more complex, interactive scenarios under closed-loop evaluation. We demonstrate that through a combination of latent feature distillation and ground-truth trajectory supervision, an efficient vision-only student model \textbf{Orion-Lite} can even surpass the performance of its massive VLA teacher, ORION. Setting a new state-of-the-art on the rigorous Bench2Drive benchmark, with a Driving Score of 80.6. Ultimately, this reveals that vision-only architectures still possess significant, untapped potential for high-performance reactive planning.
\end{abstract}

\section{Introduction}

\begin{figure}[htbp]
  \centering
  \includegraphics[width=\linewidth]{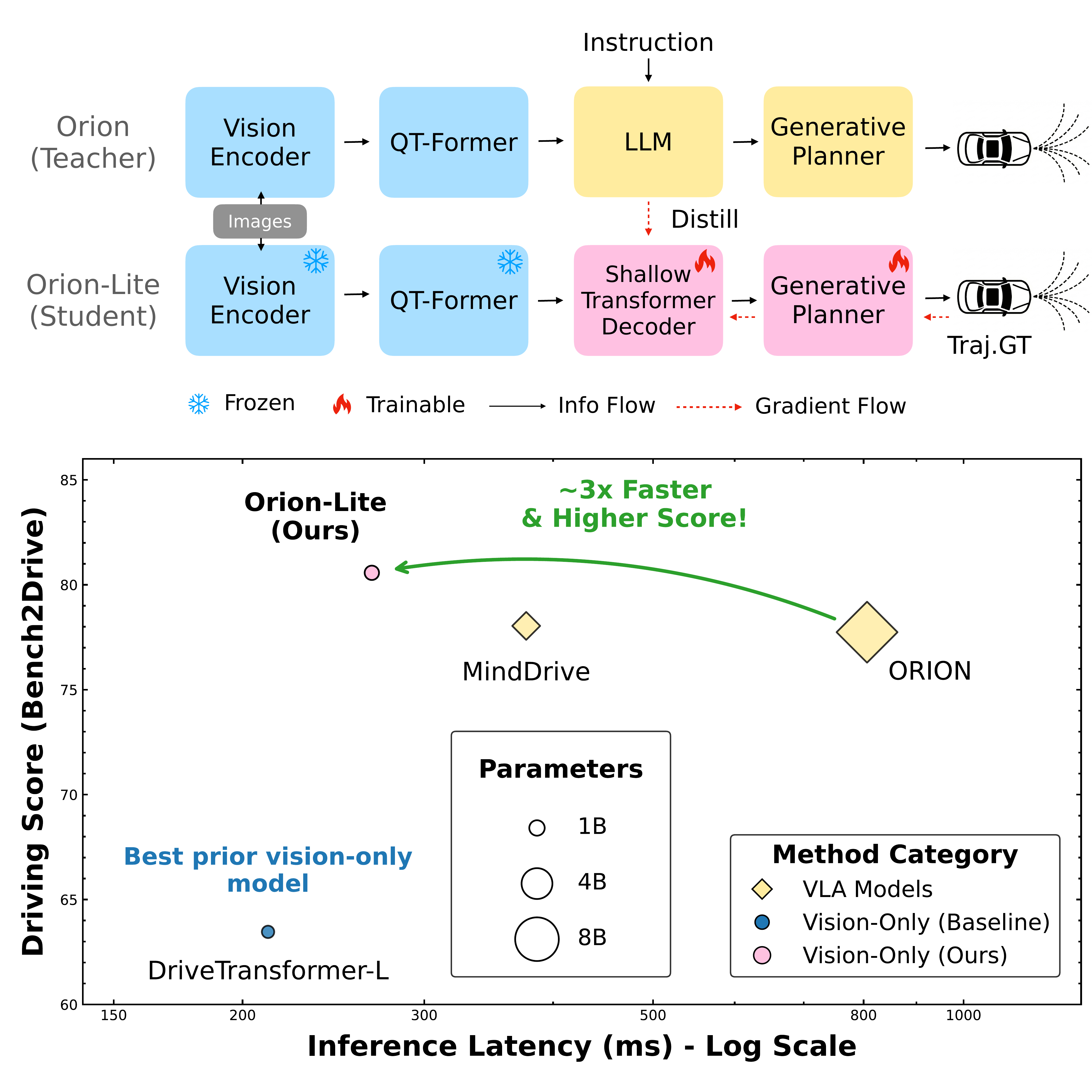}
  \caption{\textbf{Overview of the proposed distillation framework.} A joint distillation and trajectory supervision strategy (top) yields a student model, \textbf{Orion-Lite}, that is $3\times$ faster than its teacher, establishing a new state-of-the-art on the closed-loop Bench2Drive benchmark (bottom).}
  \label{fig:eye_catcher}
\end{figure}

Recently, Vision-Language Models (VLMs) and Vision-Language-Action (VLA) architectures \cite{shao2024lmdrive,tian2024drivevlm,wang2025omnidrive,xing2025openemma,xu2024vlm,fu2025orion} have emerged as a dominant paradigm in autonomous driving research. By integrating Large Language Models (LLMs) with vision encoders and aligning them with through visual question-answering (VQA), these methods can leverage the rich world knowledge embedded in LLMs. This integration of LLM modules introduces explicit causal reasoning into VLA driving models, allowing them to better optimize driving trajectories in complex, interactive scenarios \cite{fu2025orion,fu2025minddrive,xiong2026unidrive,renz2025simlingo}.

Consequently, VLA models currently achieve state-of-the-art performance across multiple autonomous driving benchmarks. However, their reliance on massive LLMs introduces severe computational bottlenecks, including prohibitive GPU memory consumption and high inference latency. While many of these architectures possess Chain-of-Thought (CoT) \cite{fu2025orion,renz2025simlingo,xiong2026unidrive} capabilities for multi-turn visual reasoning, their practical closed-loop deployment typically relies on ``direct" modes to mitigate latency. In this direct mode, intermediate reasoning steps are bypassed entirely, and the LLM is prompted with a static, pre-defined instruction template to generate latent waypoints \cite{wang2025omnidrive, fu2025orion,fu2025minddrive,xiong2026unidrive}. This essentially renders the LLM as a feature extractor.

While recent works have explored knowledge distillation to mitigate these bottlenecks, the efficacy of distilling LLMs for more challenging closed-loop driving scenarios remains an open question. For instance, DiMA \cite{hegde2025distilling} demonstrates improved inference speed by distilling LLM knowledge into a vision-only model, but its evaluation is strictly limited to open-loop metrics. Similarly, VERDI \cite{feng2025verdi} performs distillation from Qwen-2.5-VL \cite{bai2025qwen3} for closed-loop evaluation on HugSim \cite{zhou2025hugsim}. However, it primarily compares its student model against older baselines like UniAD \cite{hu2023planning} rather than against its own teacher model. Consequently, how much performance a student model can maintain relative to its teacher in more realistic, interactive environments remains underexplored. This brings us to our core research question: \textit{how can the ``reasoning" capabilities of an LLM inside a VLA be efficiently distilled without suffering a performance gap in challenging, closed-loop scenarios?}

To answer this question, we select Bench2Drive \cite{jia2024bench2drive} as our rigorous closed-loop evaluation environment. Bench2Drive is the first benchmark comprehensively designed to assess an end-to-end autonomous driving (E2E-AD) system's multi-ability performance in a closed-loop manner, introducing 44 interactive scenarios (e.g., cut-ins, overtaking, detours), 23 weather conditions, and 12 distinct towns. Historically, evaluating E2E-AD methods relied on open-loop datasets (e.g., nuScenes \cite{caesar2020nuscenes}) using L2 displacement errors and collision rates, which often fail to reflect actual driving performance \cite{jia2024bench2drive}. In fact, the average open-loop box collision rate on Bench2Drive is $4.5\times$  higher than on nuScenes (using UniAD, VAD, and ORION as baselines \cite{fu2025orion,xiong2026unidrive}), highlighting the dataset's inherent complexity. Conversely, existing closed-loop protocols (e.g., Town05Long and Longest6\cite{chitta2022transfuser,prakash2021multi}) typically rely on a small set of fixed routes, where the standard driving score exhibits high variance due to unsmoothed metric functions and route randomness. Bench2Drive bridges this gap by having more environments and scenarios, offering a stable, highly interactive evaluation standard.

Focusing on this challenging benchmark, we take the currently publicly available state-of-the-art VLA model, ORION \cite{fu2025orion}, and investigate the effect of different distillation strategies on our proposed student model, \textbf{Orion-Lite}, which replaces the heavy LLM with a lightweight transformer decoder. We find that, when employing a synergistic combination of latent distillation and ground-truth trajectory supervision, Orion-Lite surprisingly surpasses its 7B-parameter ORION teacher. Not relying on an LLM, Orion-Lite accelerates the reasoning module's inference by $150\times$, tripling the overall system speed. Moreover, it reduces total GPU memory usage from 31 GB to 8 GB, making it more suitable for actual deployment. Furthermore, qualitative analysis reveals that our distilled model exhibits superior robustness in complex edge cases where the original ORION model hesitates or fails. This lightweight, vision-only architecture achieves new state-of-the-art performance on the standard Bench2Drive benchmark, outperforming its teacher VLA-based ORION \cite{fu2025orion}, the online Reinforcement Learning based MindDrive \cite{fu2025minddrive}, and the World Model based UniDrive-WM \cite{xiong2026unidrive}.

While we do not introduce a fundamentally novel distillation algorithm, the core contribution of our work lies in the empirical demonstration that a standard distillation loss, combined with ground-truth trajectory supervision, allows a highly compressed student model to surpass its teacher's performance ceiling for challenging scenarios and under closed-loop evaluation. This suggests that the causal reasoning capability required for autonomous driving does not strictly need to manifest through a massive LLM during inference, positioning efficient visual reasoning as a highly promising direction for future research. In this work, we report these empirical findings and perform ablation studies to explore this phenomenon. A more exhaustive analysis of the underlying mechanisms is deferred to future work.

In summary, our main contributions are twofold:

\begin{itemize}
    \item \textbf{Architectural Simplification:} we demonstrate that a compact and lightweight transformer decoder can replace a 7B-parameter LLM in a VLA driving model without compromising performance on current challenging closed-loop benchmarks. This heavily suggests that for standard, reactive E2E trajectory planning, massive LLMs may not be strictly necessary for inference.
    \item \textbf{State-of-the-Art Vision-Only Model:} we show that when our shallow student model is trained jointly with feature distillation and ground-truth supervision, it effectively outperforms the teacher's driving capabilities. Our method achieves new state-of-the-art results on the Bench2Drive closed-loop evaluation, outperforming competing VLA, RL, and WM methods using only a fraction of the computational resources.
\end{itemize}
\vspace{2mm}
\noindent\textbf{Open Source:} To facilitate reproducibility and support future research in efficient autonomous driving, all code, model weights, and evaluation scripts will be made publicly available upon acceptance.

\section{Related Work}

\subsection{End-to-End Autonomous Driving}
End-to-end (E2E) autonomous driving frameworks effectively map raw sensor inputs directly to planning trajectories or control signals. UniAD \cite{hu2023planning} pioneered this direction by unifying perception, prediction, and planning into a single framework. To enhance safety, VAD \cite{jiang2023vad} incorporated vectorized planning constraints, while VADv2 \cite{chen2024vadv2} transitioned to a probabilistic paradigm. Recently, approaches like DiffusionDrive \cite{liao2025diffusiondrive} have been employed to capture multimodal trajectory distributions, and methods such as LAW \cite{li2024enhancing} and WoTE \cite{li2025end} integrate next-frame prediction as an implicit world model to enhance the vision encoder's spatial-temporal representations. Despite these advancements, many E2E methods still suffer from error accumulation in closed-loop scenarios \cite{zhai2023rethinking}. To address this, DriveTransformer \cite{jia2025drivetransformer} proposed a unified transformer framework processing perception and planning in parallel, achieving strong results on Bench2Drive. Our work builds upon this vision-only E2E paradigm but diverges significantly: we demonstrate that a high-performance vision-only model can be derived by distilling the latent cognitive capabilities of a complex VLA teacher into a drastically lighter decoder, indicating that current vision-only architectures still possess significant, untapped potential.

\subsection{Vision-Language Models}
The integration of LLMs into autonomous driving has led to the emergence of VLA models. EMMA \cite{hwang2024emma} utilizes Gemini \cite{team2023gemini} to generate future trajectories natively as text tokens. OpenEMMA \cite{xing2025openemma} extends this to open-source VLMs but relies on auxiliary 3D modules \cite{mousavian20173d}. DriveVLM \cite{tian2024drivevlm} adopts a dual-system approach for trajectory refinement. Recently, Alpamayo-R1 \cite{wang2025alpamayo} demonstrated that reinforcement learning post-training can further improve reasoning quality and reasoning-action consistency.
ORION \cite{fu2025orion} and OmniDrive \cite{wang2025omnidrive} explore using LLMs to condition generative planners. More recent works incorporate online Reinforcement Learning (MindDrive \cite{fu2025minddrive}) and World Models (UniDrive-WM \cite{xiong2026unidrive}). Crucially, in architectures like ORION and MindDrive, the final driving trajectory is decoded directly from the LLM's latent hidden states rather than its textual output. This architectural trait effectively renders the LLM as an overparameterized feature extractor. Our work capitalizes on this insight, challenging the necessity of the massive LLM during inference for standard E2E reactive planning tasks. 

\subsection{Knowledge Distillation}
Knowledge distillation (KD) transfers capabilities from heavy teacher models to lightweight student models~\cite{hinton2015distilling,beyer2022knowledge}. DriveAdapter \cite{jia2023driveadapter}, Hydra-MDP \cite{li2024hydra}, and Hydra-MDP++ \cite{li2025hydra} distill knowledge from a heavy vision-centric teacher to a more efficient vision-only student model. Distinct from these methods, our work focuses specifically on distilling the reasoning capabilities of the LLM within a VLA model into a vision-only student.

More directly relevant to our paradigm are VERDI \cite{feng2025verdi} and DiMA \cite{hegde2025distilling}, which distill VLM knowledge into vision models. VERDI aligns the VLM's text output with the vision model's predictions using complex progressive feature projectors. DiMA explores distilling VLM features via KL-divergence but limits its evaluation to open-loop metrics. We advance this research by directly distilling the continuous latent LLM features using simple $\mathcal{L}_1$ regression, avoiding auxiliary text encoders, offline rule-based experts, and suboptimal distributional metrics. Furthermore, we validate our framework in highly complex, realistic closed-loop evaluations. 

\section{Method}
In this section, we detail our proposed knowledge distillation framework, designed to effectively compress the massive LLM inside a VLA driving model without sacrificing performance in complex, closed-loop scenarios. The overall pipeline is illustrated in Figure \ref{fig:eye_catcher}. Our framework utilizes the state-of-the-art ORION \cite{fu2025orion} as the teacher model. To create our highly efficient, vision-only end-to-end model, we introduce a lightweight distillation module (Section \ref{sec:distillation_module}) that transfers the latent reasoning representations from the teacher LLM to a shallow transformer decoder. To avoid computational redundancy during distillation, we utilize the teacher's intermediate state embeddings (Section \ref{sec:state_embedding}) to represent the dense visual and contextual information.

\subsection{Preliminaries}

ORION is a fully map-free E2E method that relies on Navigation Commands (NC) as trajectory conditions, utilizing EVA-02-L \cite{fang2024eva} as the vision encoder, Vicuna-v1.5 \cite{zheng2023judging} as the LLM, and a VAE-based generative planner \cite{kingma2013auto}. 

ORION takes multi-view camera streams as input, employing a vision encoder to extract spatial features. High-level driving commands are vectorized and fused with these visual representations. To ensure robust feature learning, an auxiliary perception head is attached to the image features and supervised via downstream perception tasks. For temporal context, a QT-Former module extracts compact embeddings from historical frames, maintaining them in a memory bank. Concurrently, text prompts are tokenized and embedded before being fused with the visual features. The resulting multimodal representations are fed into an LLM, which serves two purposes: it either generates intermediate latent features for motion planning or it performs auto-regressive token generation to answer visual queries and explicit reasoning. Ultimately, the intermediate latent features are processed by a VAE-based generative planner to produce the final trajectory prediction.

\subsection{State Embedding Extraction}
\label{sec:state_embedding}

Following the ORION architecture \cite{fu2025orion}, we first extract multi-view image features $F_m$ from the frozen vision encoder. The QT-Former, a query-based temporal module, employs learnable scene queries $Q_s \in \mathbb{R}^{N_s \times C_q}$ and perception queries $Q_p \in \mathbb{R}^{N_p \times C_q}$, where $N_s$ and $N_p$ denote the number of queries and $C_q$ represents the channel dimension. 

These queries exchange information via self-attention and subsequently interact with the image features $F_m$ through cross-attention. The perception queries are then routed to task-specific heads for object detection, traffic state recognition, and dynamic agent motion prediction. To efficiently aggregate historical context, ORION utilizes history queries $Q_h \in \mathbb{R}^{N_h \times C_q}$ alongside a long-term memory bank $M \in \mathbb{R}^{(N_h \times n) \times C_q}$. 

The final concatenated features comprising the map vision embedding, ego-vehicle status, and the driving command serve as the input tokens $T_c \in \mathbb{R}^{N_c \times C_c}$ for our student model, where $N_c$ is the sequence length and $C_c$ is the channel dimension. In the teacher model, the LLM processes $T_c$ alongside tokenized text prompts to output the latent planning tokens $T_p \in \mathbb{R}^{1 \times C_p}$. For our student framework, we discard the text prompts entirely to achieve a vision-only architecture, utilizing the teacher's $T_p$ as the pseudo-ground truth distillation target.

\subsection{Lightweight Distillation Module}
\label{sec:distillation_module}
Our student module replaces the massive 7B-parameter LLM with a highly efficient transformer-based architecture. This module consists of an input projection layer, a learnable planning query, a shallow standard transformer decoder, and an output projection layer. 

The input projection, implemented as a linear layer followed by layer normalization, compresses the input token channels from $C_c$ to a hidden dimension $C_h$, yielding the compressed tokens $T_{cc} \in \mathbb{R}^{N_c \times C_h}$. To emulate the generative planning mechanism of the LLM, we initialize a learnable planning query $Q_{plan} \in \mathbb{R}^{1 \times C_h}$. Through the cross-attention mechanism within the 6-layer transformer decoder, $Q_{plan}$ (Query) attends to the dense compressed tokens $T_{cc}$ (Key/Value), extracting the critical spatio-temporal features necessary for trajectory generation. Finally, the output projection layer maps the decoder's output back to the original planning token dimension $C_p$, perfectly aligning the student's output space with the teacher's generative planner. This bottleneck design effectively retains essential planning information to manage complex scenarios while drastically reducing computational overhead.

\subsection{Training Objectives}
\label{sec:training_objectives}

\noindent\textbf{Feature Mimic Loss.} Let $T_{student} \in \mathbb{R}^{1 \times C_p}$ denote the final projected output of the student decoder. Using the teacher's generated planning tokens $T_p$ as the target, we apply an $\mathcal{L}_1$ regression loss, $\mathcal{L}_{mimic}$, to minimize the representational divergence. The pure distillation loss over a batch size $B$ is formulated as:
\begin{equation}
\label{eq:loss_naive}
\mathcal{L}_{mimic} = \frac{1}{B \cdot C_p} \sum_{b=1}^{B} \left\| T_{student}^{(b)} - T_{p}^{(b)} \right\|_1
\end{equation}

\noindent\textbf{Joint Distillation and E2E Supervision.} Rather than relying on the LLM's latent features only, our training phase jointly optimizes the newly initialized transformer decoder alongside the pre-trained VAE generative planner. Throughout this process, the pre-trained vision encoder and QT-Former are kept strictly frozen with weights initialized from ORION. Because the vision encoder already captures robust spatial-temporal representations, we focus the gradient updates entirely on the lightweight student decoder and the ensuing planner. 

Following standard E2E formulations \cite{jiang2023vad}, we incorporate environmental feedback via collision loss $\mathcal{L}_{col}$ and boundary loss $\mathcal{L}_{bd}$. Furthermore, we apply an $\mathcal{L}_1$ regression loss $\mathcal{L}_{reg}$ for deterministic trajectory prediction. To properly align the reasoning and action spaces within the generative planner, we retain the Kullback--Leibler divergence loss $\mathcal{L}_{vae}$ adopted by the ORION teacher.
To clearly separate supervision signals, we decompose the overall objective into two components: (i) a ground-truth supervision term that aggregates all driving-related penalties, and (ii) a distillation term that regularizes the student with the teacher’s latent distribution. Specifically, we define:
\begin{equation}
\label{eq:loss_gt}
\mathcal{L}_{GT} =
\mathcal{L}_{col} + 
\lambda_{bd}\mathcal{L}_{bd} +
\lambda_{reg}\mathcal{L}_{reg} +
\lambda_{vae}\mathcal{L}_{vae} 
\end{equation}

In practice, we set $\lambda_{bd} = \lambda_{reg} = \lambda_{vae} = 3$.
The final training objective is then expressed as the sum of the GT supervision and distillation terms:

\begin{equation}
\label{eq:loss_total}
\mathcal{L}_{total} = \mathcal{L}_{GT} + \mathcal{L}_{mimic}
\end{equation} 

This combination of ground truth and distillation losses ensures that the student model effectively distills knowledge from the ORION teacher, while retaining the foundational driving skills learned from the recorded trajectories.

\begin{figure}[htbp]
  \centering
  \includegraphics[width=\linewidth]{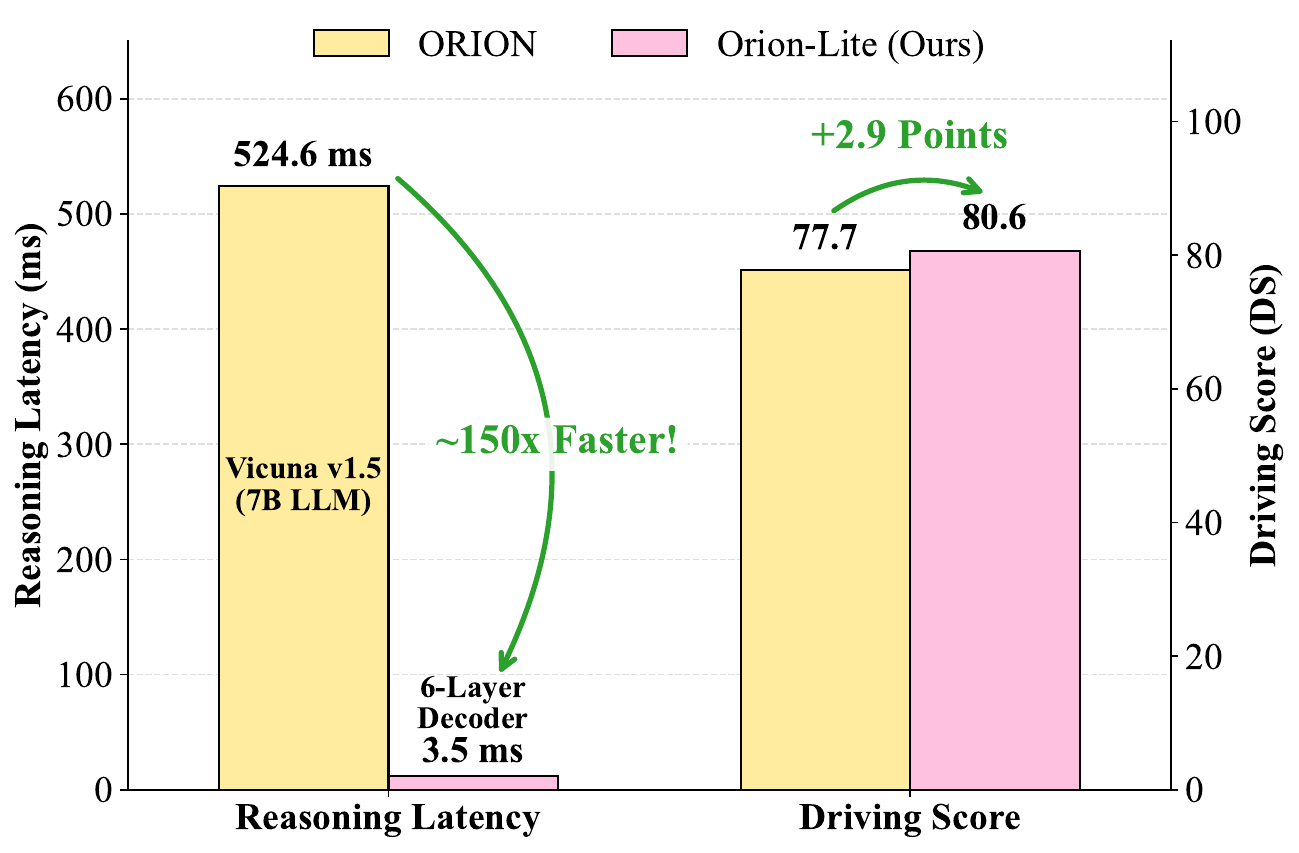}
  \caption{\textbf{Latency and Driving Score Comparison.} Our distilled framework demonstrates a massive reduction in inference latency compared to the teacher model while improving the overall Driving Score. Latency is measured by the averaged inference step-time on CARLA evaluated on an A6000 GPU.}
  \label{fig:latency and Driving score}
\end{figure}

\begin{table*}[t]
\centering
\resizebox{\textwidth}{!}{%
\begin{tabular}{l c c c >{\columncolor{colgray}}c >{\columncolor{colgray}}c c c >{\color{textgray}}c c}
\toprule
\multirow{2}{*}{\textbf{Method}} & \multirow{2}{*}{\textbf{Reference}} & \multirow{2}{*}{\textbf{Condition}} & \multirow{2}{*}{\textbf{Modality}} & \multicolumn{4}{c}{\textbf{Closed-loop Metrics}} & \textcolor{textgray}{\textbf{Open-loop Metric}} & \multirow{2}{*}{\textbf{Latency (ms)} $\downarrow$} \\
\cmidrule{5-8} \cmidrule{9-9}
~& & & & \textbf{DS} $\uparrow$ & \textbf{SR (\%)} $\uparrow$ & \textbf{Efficiency} $\uparrow$ & \textbf{Comfortness} $\uparrow$ & \textcolor{textgray}{\textbf{Avg. L2} $\downarrow$} & \\
\midrule
TCP* \cite{wu2022trajectory} & NeurIPS 22 & TP & C & 40.7 & 15.0 & 54.3 & 47.8 & 1.70 & 83 \\
TCP-ctrl* & NeurIPS 22 & TP & C & 30.5 & 7.3 & 56.0 & 51.5 & - & 83 \\
TCP-traj* & NeurIPS 22 & TP & C & 59.9 & 30.0 & 76.5 & 18.1 & 1.70 & 83 \\
TCP-traj w/o distillation & NeurIPS 22 & TP & C & 49.3 & 20.5 & 78.8 & 23.0 & 1.96 & 83 \\
ThinkTwice* \cite{jia2023think} & CVPR 23 & TP & C & 62.4 & 31.2 & 69.3 & 16.2 & 0.95 & 762 \\
DriveAdapter* \cite{jia2023driveadapter} & ICCV 23 & TP & C\&L & 64.2 & 33.1 & 70.2 & 16.0 & 1.01 & 931 \\
\midrule
AD-MLP \cite{zhai2023rethinking} & arXiv 23 & NC & C & 18.1 & 0.0 & 48.5 & 22.6 & 3.64 & 3 \\
UniAD-Tiny \cite{hu2023planning} & CVPR 23 & NC & C & 40.7 & 13.2 & 123.9 & 47.0 & 0.80 & 420 \\
UniAD-Base \cite{hu2023planning} & CVPR 23 & NC & C & 45.8 & 16.4 & 129.2 & 43.6 & 0.73 & 663 \\
VAD \cite{jiang2023vad} & ICCV 23 & NC & C & 42.4 & 15.0 & 157.9 & 46.0 & 0.91 & 278 \\
GenAD \cite{zheng2024genad} & ECCV 24 & NC & C & 44.8 & 15.9 & - & - & - & 121 \\
MomAD \cite{song2025don} & CVPR 25 & NC & C & 44.5 & 16.7 & 170.2 & 48.6 & 0.87 & 242 \\
DriveTransformer-Large \cite{jia2025drivetransformer} & ICLR 25 & NC & C & 63.5 & 35.0 & 100.6 & 20.8 & 0.62 & 212 \\
MindDrive \cite{fu2025minddrive} & arXiv 25 & NC & C & 78.0 & 55.1 & - & - & - & 377 \\
UniDrive-WM \cite{xiong2026unidrive} & arXiv 26 & NC & C & \underline{79.2} & \textbf{56.4} & 158.4 & 28.0 & 0.64 & - \\
\textcolor{gray}{SimLingo$^\dagger$} \cite{renz2025simlingo} & \textcolor{gray}{CVPR 25} & \textcolor{gray}{NC} & \textcolor{gray}{C} & \textcolor{gray}{85.9} & \textcolor{gray}{66.8} & \textcolor{gray}{244.2} & \textcolor{gray}{25.5} & \textcolor{gray}{-} & \textcolor{gray}{4139} \\
\midrule
ORION (0.5B) \cite{fu2025minddrive} & ICCV 25 & NC & C & 72.9 & 45.8 & - & - & - & - \\
ORION (7B Teacher) \cite{fu2025orion} & ICCV 25 & NC & C & 77.7 & 54.6 & 151.5 & 17.4 & 0.68 & 806 \\
\rowcolor{orionblue} Orion-Lite (\textbf{0.1B Ours}) & - & NC & C & \phantom{\scriptsize \color{oriongreen}(+2.9)} \textbf{80.6} {\scriptsize \color{oriongreen}(+2.9)} & \phantom{\scriptsize \color{oriongreen}(+0.9)} \underline{55.5} {\scriptsize \color{oriongreen}(+0.9)} & 157.7 & 10.3 & \textcolor{textgray}{0.79} & 267 \\
\bottomrule
\end{tabular}%
}
\caption{\textbf{Closed-loop and Open-loop Results of E2E-AD Methods in Bench2Drive.} All models are trained on the standard base set (1K clips), except for SimLingo$^\dagger$, which utilizes extended external training data (e.g., 3.1M samples). C/L refers to camera/LiDAR. Avg. L2 is averaged over predictions in 2 seconds under 2Hz. Latency is measured by the average inference step-time during CARLA evaluation on an A6000 GPU. * denotes expert feature distillation. NC: navigation command, TP: target point, DS: Driving Score, SR: Success Rate. Note: 0.5B and 7B indicate the parameters of the LLM module alone, rather than the full model's parameter count.}
\label{tab:bench2drive_results}
\end{table*}

\begin{table*}[t]
\centering
\resizebox{\textwidth}{!}{%
\begin{tabular}{l c c c c c c c c >{\columncolor{colgray}}c}
\toprule
\multirow{2}{*}{\textbf{Method}} & \multirow{2}{*}{\textbf{Reference}} & \multirow{2}{*}{\textbf{Condition}} & \multirow{2}{*}{\textbf{Modality}} & \multicolumn{6}{c}{Ability (\%) $\uparrow$} \\
\cmidrule{5-10}
~& & & & Merging & Overtaking & Emergency Brake & Give Way & Traffic Sign & Mean \\
\midrule
TCP* \cite{wu2022trajectory} & NeurIPS 22 & TP & C & 16.2 & 20.0 & 20.0 & 10.0 & 7.0 & 14.6 \\
TCP-ctrl* & NeurIPS 22 & TP & C & 10.3 & 4.4 & 10.0 & 10.0 & 6.5 & 8.2 \\
TCP-traj* & NeurIPS 22 & TP & C & 8.9 & 24.3 & 51.7 & \underline{40.0} & 46.3 & 34.2 \\
TCP-traj w/o distillation & NeurIPS 22 & TP & C & 17.1 & 6.7 & 40.0 & \textbf{50.0} & 28.7 & 28.5 \\
ThinkTwice* \cite{jia2023think} & CVPR 23 & TP & C & 27.4 & 18.4 & 35.8 & \textbf{50.0} & 54.2 & 37.2 \\
DriveAdapter* \cite{jia2023driveadapter} & ICCV 23 & TP & C\&L & 28.8 & 26.4 & 48.8 & \textbf{50.0} & 56.4 & 42.1 \\
\midrule
AD-MLP \cite{zhai2023rethinking} & arXiv 23 & NC & C & 0.0 & 0.0 & 0.0 & 0.0 & 4.35 & 0.87 \\
UniAD-Tiny \cite{hu2023planning} & CVPR 23 & NC & C & 8.9 & 9.3 & 20.0 & 20.0 & 15.4 & 14.7 \\
UniAD-Base \cite{hu2023planning} & CVPR 23 & NC & C & 14.1 & 17.8 & 21.7 & 10.0 & 14.2 & 15.6 \\
VAD \cite{jiang2023vad} & ICCV 23 & NC & C & 8.1 & 24.4 & 18.6 & 20.0 & 19.2 & 18.1 \\
DriveTransformer-Large \cite{jia2025drivetransformer} & ICLR 25 & NC & C & 17.6 & 35.0 & 48.4 & \underline{40.0} & 52.1 & 38.6 \\
MindDrive \cite{fu2025minddrive} & arXiv 25 & NC & C & \textbf{32.9} & \textbf{75.8} & 68.3 & \textbf{50.0} & 57.9 & 56.9 \\
UniDrive-WM \cite{xiong2026unidrive} & arXiv 26 & NC & C & \underline{29.8} & 74.0 & \textbf{79.8} & \underline{40.0} & \textbf{71.3} & \underline{59.0} \\
\midrule
ORION (0.5B) \cite{fu2025minddrive} & ICCV 25 & NC & C & 26.3 & 62.2 & 55.6 & \textbf{50.0} & 63.3 & 51.4 \\
ORION (7B Teacher) \cite{fu2025orion} & ICCV 25 & NC & C & 25.0 & 71.1 & \underline{78.3} & 30.0 & 69.2 & 54.7 \\
\rowcolor{orionblue} Orion-Lite (\textbf{0.1B Ours}) & - & NC & C & 28.8 & \underline{75.6} & \underline{78.3} & \textbf{50.0} & \underline{70.0} & \phantom{{\scriptsize \color{oriongreen}(+5.8)}} \textbf{60.5} {\scriptsize \color{oriongreen}(+5.8)} \\
\bottomrule
\end{tabular}%
}
\caption{\textbf{Multi-Ability Results of E2E-AD Methods under base set.} * denote expert feature distillation. C/L refers to camera/LiDAR. NC: navigation command, TP: target point.}
\label{tab:multi_ability_results}
\end{table*}

\begin{figure*}[t]
  \centering
  \includegraphics[width=\textwidth]{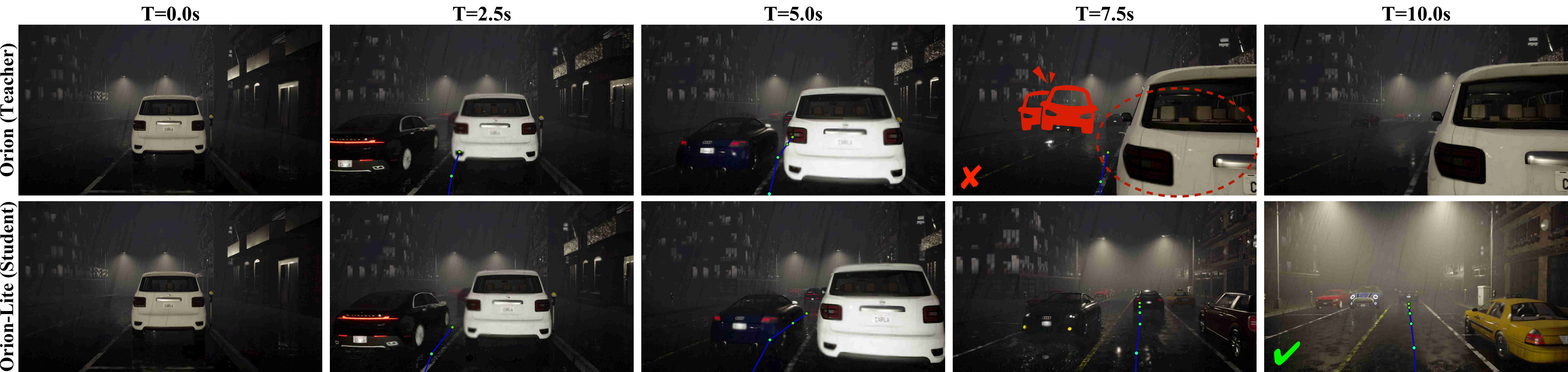}\par\medskip
  \includegraphics[width=\textwidth]{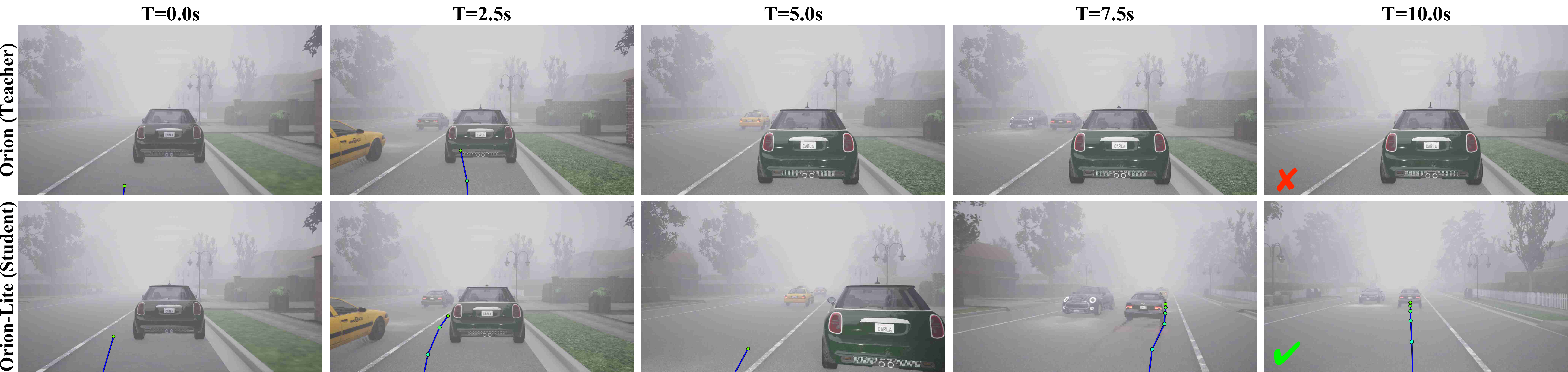}
  \caption{\textbf{Qualitative comparison in interactive scenarios.} Top rows: rollouts from the Orion teacher model. Bottom rows: rollouts from our student model. Sequences are visualized at 5-frame intervals, with overlaid points indicating predicted future trajectories. Green checks ({\color{green}$\checkmark$}) denote successful, intervention-free maneuvers, while red crosses ({\color{red}$\times$}) indicate task failures. While Orion frequently hesitates or fails to safely merge or overtake behind obstacles, our student model demonstrates robust spatial awareness, successfully executing smooth and decisive maneuvers.}
  \label{fig:Qualitative Results}
\end{figure*}

\begin{table}[t]
    \centering
    \setlength{\tabcolsep}{4pt}
    \resizebox{\columnwidth}{!}{%
        \begin{tabular}{@{}cc cc cccccc@{}}
            \toprule
            \multirow{2}{*}{\textbf{Mimic Loss}} & \multirow{2}{*}{\textbf{Traj. GT}} & \multicolumn{2}{c}{\textbf{Closed-loop}} & \multicolumn{6}{c}{\textbf{Ability (\%)} $\uparrow$} \\
            \cmidrule(lr){3-4} \cmidrule(l){5-10}
             & & \textbf{DS} $\uparrow$ & \textbf{SR} $\uparrow$ & M & O & EB & GW & TS & Mean \\
            \midrule
            - & \checkmark & 73.9 & 50.0 & 26.3 & 66.7 & 71.7 & 10.0 & 63.7 & 47.7 \\
            \checkmark & - & 76.0 & 50.7 & 26.6 & 66.7 & 68.3 & 40.0 & 64.7 & 53.3 \\
            \checkmark & \checkmark & \textbf{80.6} & \textbf{55.5} & \textbf{28.8} & \textbf{75.6} & \textbf{78.3} & \textbf{50.0} & \textbf{70.0} & \textbf{60.5} \\
            \bottomrule
        \end{tabular}%
    }
    \caption{\textbf{Impact of supervision.} We evaluate the model's performance on closed-loop metrics and specific driving abilities. \textbf{M}: Merging, \textbf{O}: Overtaking, \textbf{EB}: Emergency Brake, \textbf{GW}: Give Way, \textbf{TS}: Traffic Sign, \textbf{DS/SR}: Driving Score/Success Rate.}
    \label{tab:ablation_mimic}
\end{table}

\begin{table}[t]
    \centering
    \setlength{\tabcolsep}{4pt}
    \resizebox{\columnwidth}{!}{%
        \begin{tabular}{@{}lc cc cccccc@{}}
            \toprule
            \multirow{2}{*}{\textbf{Setting}} & \multirow{2}{*}{\textbf{Epochs}} & \multicolumn{2}{c}{\textbf{Closed-loop}} & \multicolumn{6}{c}{\textbf{Ability (\%)} $\uparrow$} \\
            \cmidrule(lr){3-4} \cmidrule(l){5-10}
             & & \textbf{DS} $\uparrow$ & \textbf{SR} $\uparrow$ & M & O & EB & GW & TS & Mean \\
            \midrule
            Orion & 18 (default) & 77.7 & 54.6 & 25.0 & 71.1 & \textbf{78.3} & 30.0 & 69.2 & 54.7 \\
            Orion & 24 & 77.1 & 50.7 & 25.3 & 57.8 & 70.0 & 40.0 & 69.0 & 52.0 \\
            Orion-Lite & 20 & \textbf{80.6} & \textbf{55.5} & \textbf{28.8} & \textbf{75.6} & \textbf{78.3} & \textbf{50.0} & \textbf{70.0} & \textbf{60.5} \\
            
            \bottomrule
        \end{tabular}%
    }
    \caption{\textbf{Impact of training duration.} Comparison between the standard Orion model, Orion with extended training duration and our Orion-Lite.}
    \label{tab:long_training}
\end{table}

\begin{table}[t]
    \centering
    \scriptsize %
    \setlength{\tabcolsep}{7.1pt} %
    \begin{tabular}{@{}l cccc cccc@{}}
        \toprule
        \multirow{2}{*}{\textbf{Loss}} & \multicolumn{4}{c}{\textbf{L2 (m)} $\downarrow$} & \multicolumn{4}{c}{\textbf{Collision (\%)} $\downarrow$} \\
        \cmidrule(lr){2-5} \cmidrule(l){6-9}
         & \textcolor{gray}{\textbf{1s}} & \textcolor{gray}{\textbf{2s}} & \textcolor{gray}{\textbf{3s}} & \textbf{Avg.} & \textcolor{gray}{\textbf{1s}} & \textcolor{gray}{\textbf{2s}} & \textcolor{gray}{\textbf{3s}} & \textbf{Avg.} \\
        \midrule
        $\mathcal{L}_1$ & \textcolor{gray}{0.32} & \textcolor{gray}{0.76} & \textcolor{gray}{1.30} & 0.79 & \textcolor{gray}{0.27} & \textcolor{gray}{0.51} & \textcolor{gray}{0.85} & 0.54 \\
        $\mathcal{L}_2$ & \textcolor{gray}{0.32} & \textcolor{gray}{0.75} & \textcolor{gray}{1.30} & 0.79 & \textcolor{gray}{0.36} & \textcolor{gray}{0.55} & \textcolor{gray}{0.90} & 0.60 \\
        KL & \textcolor{gray}{0.32} & \textcolor{gray}{0.75} & \textcolor{gray}{1.30} & 0.79 & \textcolor{gray}{0.34} & \textcolor{gray}{0.63} & \textcolor{gray}{0.91} & 0.63 \\
        Huber & \textcolor{gray}{0.31} & \textcolor{gray}{0.70} & \textcolor{gray}{1.24} & 0.75 & \textcolor{gray}{0.41} & \textcolor{gray}{0.64} & \textcolor{gray}{1.04} & 0.70 \\
        \bottomrule
    \end{tabular}
    \caption{\textbf{Impact of distance metrics.} We evaluate the impact of different distance metrics for distillation.}
    \label{tab:ablation_loss_collision}
\end{table}

\begin{table}[t]
    \centering
    \setlength{\tabcolsep}{4pt}
    \resizebox{\columnwidth}{!}{%
        \begin{tabular}{@{}lc cc cccccc@{}}
            \toprule
            \multirow{2}{*}{\textbf{Encoder Init.}} & \multirow{2}{*}{\textbf{Frozen}} & \multicolumn{2}{c}{\textbf{Closed-loop}} & \multicolumn{6}{c}{\textbf{Ability (\%)} $\uparrow$} \\
            \cmidrule(lr){3-4} \cmidrule(l){5-10}
             & & \textbf{DS} $\uparrow$ & \textbf{SR} $\uparrow$ & M & O & EB & GW & TS & Mean \\
            \midrule
            EVA-02-L & \checkmark & 54.3 & 21.8 & 11.3 & 33.3 & 28.3 & \textbf{30.0} & 41.1 & 28.8 \\
            EVA-02-L &  & 72.4 & 47.3 & 23.8 & 53.3 & \textbf{70.0} & \textbf{30.0} & \textbf{67.9} & 49.0 \\
            Orion & \checkmark & \textbf{77.4} & \textbf{51.8} & \textbf{30.0} & \textbf{66.7} & 68.3 & \textbf{30.0} & 65.8 & \textbf{52.2} \\
            \bottomrule
        \end{tabular}%
    }
\caption{\textbf{Ablation on vision encoder initialization.} We evaluate driving performance using different pre-trained weights.}
    \label{tab:ablation_encoder}
\end{table}

\begin{figure}[htbp]
  \centering
  \includegraphics[width=\linewidth]{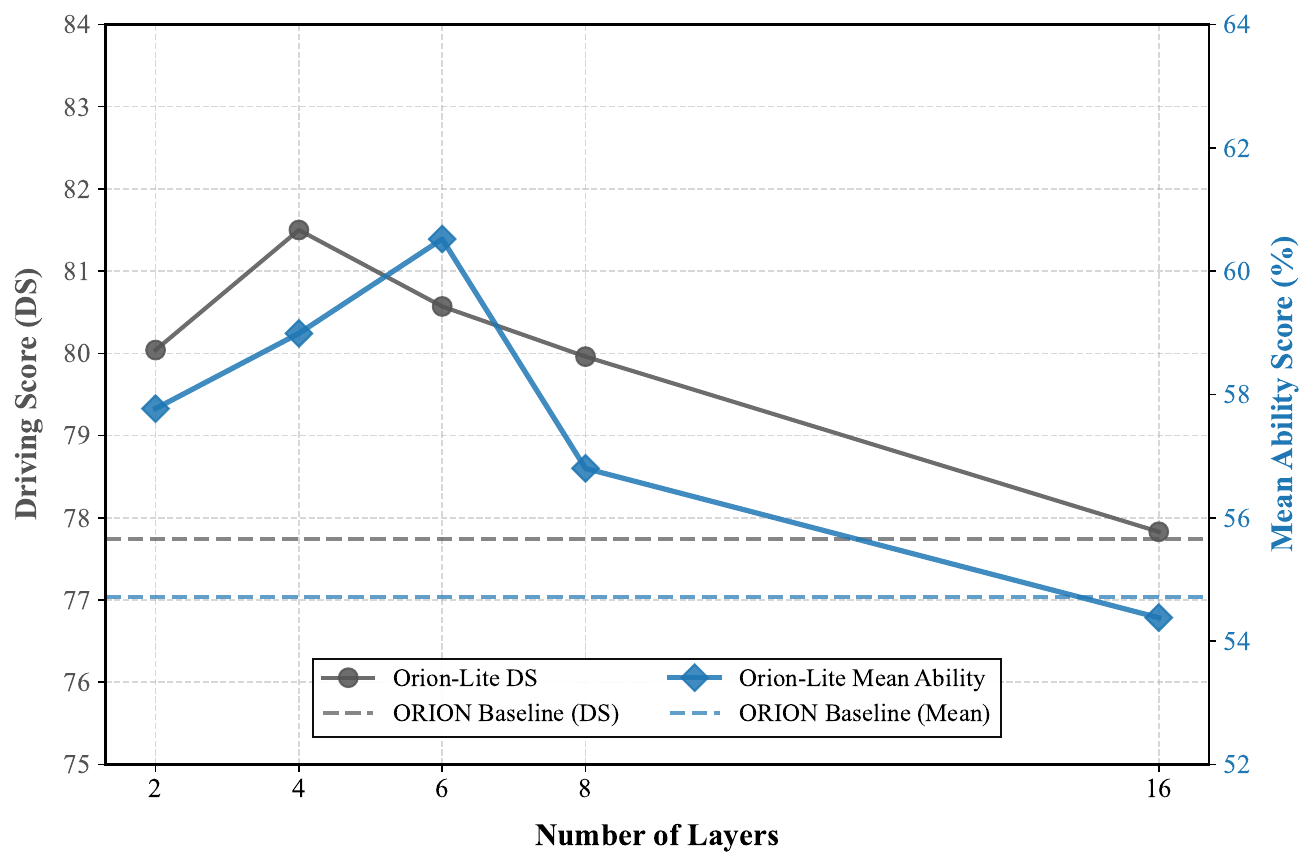}
  \caption{\textbf{Impact of Decoder Depth.} Driving Score and mean Multi-ability Score across varying numbers of transformer decoder layers.}
  \label{fig:ablation layer number of the decoder layer}
\end{figure}

\section{Experiments}

\subsection{Dataset}
We train and evaluate our models utilizing the Bench2Drive dataset \cite{jia2024bench2drive}, which employs CARLA V2 \cite{dosovitskiy2017carla} as its closed-loop evaluation protocol for E2E autonomous driving. The official training set contains 1,000 annotated clips, each comprising multi-view camera data (6 cameras), 5 radars, and 1 LiDAR sweep. Radar and LiDAR sweeps are discarded, as our model leverages RGB cameras only. Each clip spans roughly 150 meters and captures a specific interactive driving scenario. For our distillation pipeline, we utilize 950 clips for training and 50 for open-loop validation. Comprehensive closed-loop evaluations are conducted on the official Bench2Drive CARLA simulator, encompassing 220 short routes across 44 complex, interactive scenarios.

\subsection{Evaluation Metrics}
Following the Bench2Drive benchmark, we report five different metrics for closed-loop evaluation: Driving Score (DS), Success Rate (SR), Efficiency, Comfortness, and Multi-Ability.
\textit{Driving Score}, standard in CARLA \cite{dosovitskiy2017carla}, is the primary metric, multiplying the route completion percentage by a penalty discount factor for any infractions (e.g., collisions, running red lights). \textit{Success Rate} measures the percentage of successfully completed routes within a designated time limit. \textit{Efficiency} and \textit{Comfortness} quantify the agent's navigational speed and kinematic smoothness, respectively. \textit{Multi-Ability} independently evaluates the model across five advanced urban driving skills (Merging, Overtaking, Emergency Braking, Giving Way, and Traffic Signs). 
For the sake of completeness, we also report open-loop validation. In this case, analogously to ST-P3 \cite{hu2022st}, we report the L2 trajectory displacement error (Avg. L2) and the bounding box collision rate.

\subsection{Implementation Details}

\textbf{Model Settings.} Unlike VLA-based frameworks \cite{fu2025orion, xiong2026unidrive}, our proposed Orion-Lite is a vision-only architecture that receives raw camera frames as input and directly yields trajectory predictions.

\noindent\textbf{Training Process.} We initialize the vision encoder, QT-Former, and VAE planner with pre-trained ORION \cite{fu2025orion} weights. The 7B LLM is replaced by our randomly initialized 6-layer decoder (reducing reasoning parameters from 7B to 0.1B). During distillation, the vision encoder and QT-Former remain frozen; only the student decoder and VAE planner are updated. All ablations adopt this setting unless otherwise specified. Models are trained for 20 epochs at $640 \times 640$ resolution using AdamW with learning rate $5 \times 10^{-5}$ and weight decay $1 \times 10^{-4}$. Training requires $\sim$20 hours on a single RTX A6000 (48GB) GPU. Closed-loop evaluations utilize the same device. Further details are provided in our open-source repository.

\section{Results}

\subsection{Main Results}
As detailed in Figure \ref{fig:latency and Driving score} and Table \ref{tab:bench2drive_results}, our distilled model yields exceptional efficiency gains while achieving superior performance with respect to the teacher model. During the inference phase, the reasoning module of our student model is 150$\times$ faster than the VLA teacher's LLM and it reduces the inference GPU memory usage from 31 GB to 8 GB. This results in a 3$\times$ decrease of the overall end-to-end system latency. On top of these massive benefits in latency and memory consumption, our distilled model surpasses the ORION teacher by +2.9 DS, +0.9 SR, and +5.8 in Mean Multi-Ability (see Table \ref{tab:multi_ability_results}). 

Evaluated under the  Bench2Drive closed-loop evaluation, our lightweight vision-only framework establishes a new state-of-the-art across closed-loop metrics, also outperforming the recent online RL and WM-based E2E-AD models\cite{fu2025minddrive,xiong2026unidrive}. This explicitly answers our core research question: \textit{through our distillation framework, the reasoning capabilities of a massive LLM can be effectively compressed into a lightweight transformer decoder without compromising and, in fact, improving performance in challenging closed-loop scenarios.} 

Figure \ref{fig:Qualitative Results} visualizes the closed-loop behaviour of our student model compared to the ORION teacher in highly interactive scenarios. The comparative rollout highlights a specific case where the ego-vehicle is navigating around dynamic obstacles. In this scenario, the ORION teacher’s generative planner frequently hesitates or fails to execute a safe merge or overtake when positioned behind an obstacle. Conversely, our student model maintains robust spatial awareness and successfully executes a smooth maneuver. 

In the subsequent ablation study, we conduct further experiments to verify whether both the mimic loss and trajectory supervision are strictly necessary for this performance leap, while also analyzing the role of the mimic loss during training and the efficacy of various mimic loss formulations.

\subsection{Ablation Study}
\textbf{The Role of Mimic Loss.} 
Table \ref{tab:ablation_mimic} isolates the impact of the mimic loss. When training the shallow decoder from scratch relying solely on trajectory ground truth ($\mathcal{L}_{GT}$ without $\mathcal{L}_{mimic}$), performance drops to 73.9 DS. Notably, this still represents $\sim$95\% of the teacher's performance, indicating that the frozen ORION vision encoder already provides a highly robust spatio-temporal foundation. Conversely, applying the mimic loss alone recovers 98\% of the teacher's performance (76.0 DS). However, when applied jointly, the model achieves SOTA performance (80.6 DS).
This shows that a synergistic combination of latent feature distillation and standard GT supervision can deliver better performance than the teacher itself, for a much lower inference footprint.

\noindent\textbf{Impact of Training Duration.} 
As shown in Table \ref{tab:long_training}, directly training the ORION teacher for extended epochs (from 18 to 24 epochs) purely on ground truth results in performance degradation (77.7 DS $\rightarrow$ 77.1 DS), likely due to overfitting to the training dataset. In contrast, our joint distillation model can be stably trained for 20 epochs to achieve 80.6 DS. We attribute this to the powerful regularizing effect of knowledge distillation. The teacher LLM's latent embeddings act as soft labels, smoothing the target distribution and preventing the student from overfitting to hard, deterministic trajectory targets.

\noindent\textbf{Decoder Depth.} 
Figure \ref{fig:ablation layer number of the decoder layer} illustrates the effect of scaling the student decoder's depth. When utilizing mimic loss, a highly compressed 4-layer model achieves a remarkable peak in Driving Score (81.5 DS). However, the 6-layer configuration achieves a superior Mean Multi-Ability score (60.5\%), demonstrating a more robust mastery across diverse, complex driving skills (e.g., merging) rather than merely optimizing the base navigation score. Consequently, we adopt the 6-layer architecture as our default Orion-Lite model. Scaling beyond 6 layers causes both DS and Mean Ability Score to steadily decline. This suggests that mapping to the LLM's latent driving intent requires relatively low representational capacity; an overly deep student network risks overfitting to the distillation task itself, thereby degrading generalization in unseen closed-loop scenarios.

\noindent\textbf{Distance Metrics for Distillation.} 
Table \ref{tab:ablation_loss_collision} evaluates various distance metrics for $\mathcal{L}_{mimic}$. Because the target planning tokens ($T_p$) are dense, continuous feature coordinates rather than discrete class logits, applying distributional metrics like KL-Divergence requires artificially normalizing the feature space (e.g. via softmax), which distorts the latent geometry. Consequently, standard Euclidean regression metrics ($\mathcal{L}_1$, $\mathcal{L}_2$) naturally outperform KL-Divergence for feature matching. Specifically, $\mathcal{L}_1$ yields the lowest collision rates. We attribute this to the fact that $\mathcal{L}_1$ regression is inherently more robust to the extreme outlier activations.  

\noindent\textbf{Vision Encoder Initialization.}
To assess whether the teacher training stage provides improved visual representations for driving, Table \ref{tab:ablation_encoder} compares different vision encoder initializations under trajectory-only supervision (i.e., without the mimic loss). We adopt Orion-Lite as the base architecture and keep all other settings unchanged unless stated otherwise.
Initializing the encoder from the Orion teacher and keeping it frozen yields the best performance, achieving 77.4 DS and 51.8 SR. In contrast, replacing it with an EVA-02-L initialization while still freezing the encoder results in a substantial drop, to 54.3 DS and 21.8 SR. This large performance gap suggests that the teacher-trained encoder already captures driving-specific visual representations that are absent in a generic initialization.
When the EVA-02-L encoder is unfrozen and fine-tuned, performance improves significantly to 72.4 DS and 47.3 SR, indicating that trajectory supervision can adapt a generic backbone to the driving domain. However, it still falls short of the frozen Orion initialization by 5.0 DS. These findings suggest that Orion’s vision encoder acquires important semantic and spatial priors during joint training with the LLM, which are not easily recovered through trajectory supervision alone.

\section{Limitations and Future Work}
\label{sec:limits}

While our student model achieves a $150\times$ speedup in the reasoning module, the heavy pre-trained vision encoder (e.g., EVA-02-L) becomes the primary computational bottleneck during overall system inference. Furthermore, our distillation pipeline fundamentally relies on the prior existence of a fully trained, computationally expensive VLA teacher model. Additionally, our empirical validation is currently focused on the Bench2Drive benchmark. Although Bench2Drive is one of the most challenging benchmarks, future research should verify these findings across diverse driving datasets and explore the optimization the vision encoder itself together with the QT-Former. It is recommended to enhance the Bench2Drive test set with extra complex long-tail, edge-case scenarios to better research the need for VLA reasoning in autonomous driving. Additionally, developing novel frameworks capable of injecting broad world knowledge from an LLM directly into a visual reasoning module without the need to curate massive, domain-specific VQA datasets presents a highly promising direction to circumvent the need for massive teacher models entirely.

\section{Conclusion}

In this work, we address the severe computational bottlenecks of deploying Vision-Language-Action (VLA) models by introducing a streamlined, highly effective knowledge distillation framework. We show how to distill the latent reasoning capabilities of a massive 7B-parameter LLM into a shallow, efficient transformer decoder without suffering a performance gap in challenging, closed-loop scenarios. In fact, through joint distillation and ground truth trajectory supervision, our vision-only student model consistently surpasses its massive VLA teacher in these highly complicated driving scenarios.

Our proposed Orion-Lite architecture drastically reduces inference latency and memory footprint while establishing a new state-of-the-art on the complex Bench2Drive benchmark. Ultimately, our findings suggest that rather than indefinitely scaling architectural parameters for inference, optimizing training paradigms and distilling latent reasoning capabilities can unlock significant, untapped potential in vision-only end-to-end autonomous driving.

{
    \small
    \bibliographystyle{ieeenat_fullname}
    \bibliography{main}
}

\end{document}